\documentclass[conference]{IEEEtran}
\IEEEoverridecommandlockouts

\usepackage{cite}
\usepackage{amsmath,amssymb,amsfonts}
\usepackage{algorithmic}
\usepackage{graphicx}
\usepackage{textcomp}
\usepackage{xcolor}
\usepackage{hyperref}
\usepackage{subcaption}
\def\BibTeX{{\rm B\kern-.05em{\sc i\kern-.025em b}\kern-.08em
    T\kern-.1667em\lower.7ex\hbox{E}\kern-.125emX}}
\begin{document}


\title{Mapping Semantic Segmentation to Point Clouds Using Structure from Motion for Forest Analysis}


\author{\IEEEauthorblockN{Francisco Raverta Capua}
\IEEEauthorblockA{\textit{Universidad de Buenos Aires.} \\ \textit{Facultad de Ciencias Exactas y Naturales.} \\
\textit{CONICET-Universidad de Buenos Aires.} \\ \textit{Instituto de Ciencias de la Computación (ICC)}\\
Buenos Aires, Argentina. \\
\href{mailto:fraverta@icc.fcen.uba.ar}{fraverta@icc.fcen.uba.ar}}
\and
\IEEEauthorblockN{Pablo De Cristóforis}
\IEEEauthorblockA{\textit{Universidad de Buenos Aires.} \\ \textit{Facultad de Ciencias Exactas y Naturales.} \\
\textit{CONICET-Universidad de Buenos Aires.} \\ \textit{Instituto de Ciencias de la Computación (ICC)}\\
Buenos Aires, Argentina. \\
\href{mailto:pdecris@dc.uba.ar}{pdecris@dc.uba.ar}}
}

\maketitle

\begin{abstract}
Although the use of remote sensing technologies for monitoring forested environments has gained increasing attention, publicly available point cloud datasets remain scarce due to the high costs, sensor requirements, and time-intensive nature of their acquisition. Moreover, as far as we are aware, there are no public annotated datasets generated through Structure From Motion (SfM) algorithms applied to imagery, which may be due to the lack of SfM algorithms that can map semantic segmentation information into an accurate point cloud, especially in a challenging environment like forests.

In this work, we present a novel pipeline for generating semantically segmented point clouds of forest environments. Using a custom-built forest simulator, we generate realistic RGB images of diverse forest scenes along with their corresponding semantic segmentation masks. These labeled images are then processed using modified open-source SfM software capable of preserving semantic information during 3D reconstruction. The resulting point clouds provide both geometric and semantic detail, offering a valuable resource for training and evaluating deep learning models aimed at segmenting real forest point clouds obtained via SfM.

\end{abstract}


\section{Introduction}

In recent years, remote sensing growth, driven by rapid development of sensor hardware and advances in machine learning techniques, made it possible to study difficult environments and, in consequence, advance the frontiers of precision forest management~\cite{murtiyoso2024}. However, even though the development of techniques such as aerial photogrammetry and terrestrial or aerial laser scanning has made possible the fast acquisition of data, there are very few publicly available datasets. 

Creating a dataset of a forest environment is still expensive, as high-end equipment is needed to survey the specified area, and it is time-consuming, as it implies manual labeling. In addition, some areas may be inaccessible or dangerous to humans, further complicating data collection. For these reasons, there are very few publicly available forest environment datasets. Furthermore, these available datasets are either images or orthomosaics taken via photogrammetry, or point clouds taken via LiDAR scanning, but, to our knowledge, there are no labeled datasets available of point clouds generated via applying Structure from Motion (SfM) algorithms to images of forestry. 

Both LiDAR and cameras made it possible to acquire three-dimensional data of the studied environment. Laser scanning, while being the most accurate method for 3D acquisition, is more expensive, heavier, and energy-consuming, and therefore cameras are often used as a cheaper, lighter, and hence a more accessible alternative. 
Moreover, labeling 2D images is considered an easier task in terms of data availability and algorithm maturity~\cite{stathopoulou2019}. 
However, due to lower accuracy in the 3D representation of the environment, there are no available datasets containing point clouds generated via SfM algorithms. 
We believe that this may be due to the lack of SfM algorithms that can map semantic segmentation information to an accurate point cloud of a challenging environment. 

In the field of precision forestry, various tools are employed to generate maps, 3D models, and orthomosaics through photogrammetry, including commercial solutions such as Pix4D~\cite{pix4d}, Agisoft~\cite{agisoft}, and DroneDeploy~\cite{dronedeploy}. Among open-source alternatives, OpenDroneMap~\cite{opendronemapsweb} stands out for its robust feature set, which supports the generation of highly accurate results.

In this work, we modify the well-known state-of-the-art OpenDroneMaps software to enable mapping semantic segmentation information from imagery to the resulting  point cloud. We believe that this approach will contribute to the three-dimensional analysis of forest environments that is typically restricted to the data acquired via laser scanning only, such as volumetric or structural parameter estimation. For this purpose, a custom-built forest simulator~\cite{raverta2024} is used to generate realistic synthetic forest scenes and the corresponding RGB imagery from a top-down view along with the semantic segmentation labels. Although a real-forest dataset with semantic segmentation can also be used for this purpose, we choose to use the synthetic images not only because it is easy to obtain several data samples from the simulator, but also because the semantic segmentation, which is automatically generated, can be considered ground-truth. 

\section{Related Work}

Nowadays, the most used open-source 3D reconstruction softwares do not have the capability to use the semantic information of the image input. However, in recent years, several works proposed ways to incorporate semantic segmentation information to Structure from Motion algorithms.~\cite{wei2024} proposed a semantically-aware reconstruction method for urban application, leveraging semantic consistency to assess a better depth estimation.~\cite{murtiyoso2022} proposed a deep-learning method to incorporate semantic segmentation into photogrammetric 3D reconstruction and tested it with building imagery. Similarly,~\cite{stathopoulou2019} trained a Convolutional Neural Network to improve the performance of 3D reconstruction, also testing the result in a building imagery.~\cite{park2021} uses a deep learning network to semantically segment images of a building, and then applies an SfM algorithm to obtain the semantic point cloud data. Additionally,~\cite{chen2018} presented a way to boost the accuracy of feature point matching using semantic segmentation information. 
Although all these works contributed significantly to the integration of semantic segmentation information to the generation of labeled point clouds obtained via SfM algorithms, they have been largely tested in environments that are relatively easy to reconstruct, and use SfM algorithms that lack some additional features necessary for more challenging unstructured environments. 
In this work, we will focus on reconstructing a forest environment, which presents a greater challenge for SfM due to its complexity and unstructured nature. Moreover, since reconstructions in such environments often rely on auxiliary features, such as ground control points, to improve accuracy, we propose modifying a state-of-the-art open-source SfM algorithm that already supports these features. Our goal is to modify and extend the OpenDroneMaps software by enabling the incorporation of semantic segmentation information into the final point cloud.

Few publicly available forest environment datasets are available. Among them,~\cite{kaijaluoto2022} presented a semantically segmented point cloud of a forest region of Evo, Finland, captured using a LiDAR, and used deep neural networks to label the point cloud into several categories;~\cite{beloiu2023} developed an image dataset set in forests from the northern part of Switzerland, and used a Faster-RCNN network to detect and classify different tree species, and~\cite{cloutier2023} presented an image dataset from a forest of Canada, and used a CNN to segment and classifying different tree species. However, to our knowledge, there are no public datasets of forest environments consisting of a point cloud generated by applying SfM algorithms to forest imagery. This may be due to the lack of SfM algorithms that can be used to obtain an accurate labeled point cloud as output, and also the lack of a large semantically segmented labeled image dataset. 

In~\cite{raverta2024}, a synthetic forest environment simulator capable of creating realistic forest scenes was developed. Using it, two point cloud datasets were developed by extracting points from the meshes of the objects placed on the generated scenes. With these datasets, several deep learning networks were trained to segment point clouds and then tested with real forest data, concluding that the use of synthetic data for training can be used for later segmenting the real forest point cloud. This simulator also has the capability of generating a survey to extract synthetic images from the created scene, including RGB and semantic data, and also has the capacity of exporting ground control point information. However, the approach of using the synthetic images to generate a point cloud and then train the networks with them has not yet been explored. In this work, we will continue this idea, presenting a novel pipeline to generate semantically segmented point clouds of forest environments. 

\section{Materials and Methods}

\subsection{Forest Simulator and Synthetic Data Generation}
In~\cite{raverta2024}, we developed a forest simulator based on the Unity engine\footnote[1]{\url{https://github.com/lrse/forest-simulator}}, which is capable of procedurally generating realistic synthetic forest scenes. This simulator allows for the fast generation of large volumes of data that can be used to train deep learning networks. As one of the main features, it can export the point cloud of the meshes of all the objects that constitute the scene. Also, it can simulate a survey and export the images of the scene taken from a top-down view and export them along with the corresponding semantic segmentation labels. Additionally, the simulator is capable of working with ground control points, which helps SfM algorithms to obtain more accurate 3D reconstructions. 

\begin{figure}[t!]
    \centering
    \begin{subfigure}[b]{0.42\textwidth}
        \includegraphics[width=\textwidth]{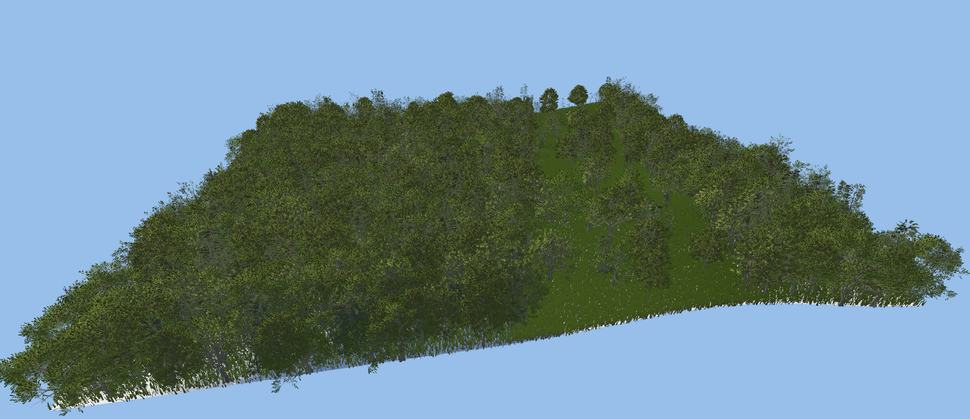}
    \end{subfigure}\\
    \vspace{1mm}
    \begin{subfigure}[b]{0.42\textwidth}
        \includegraphics[width=\textwidth]{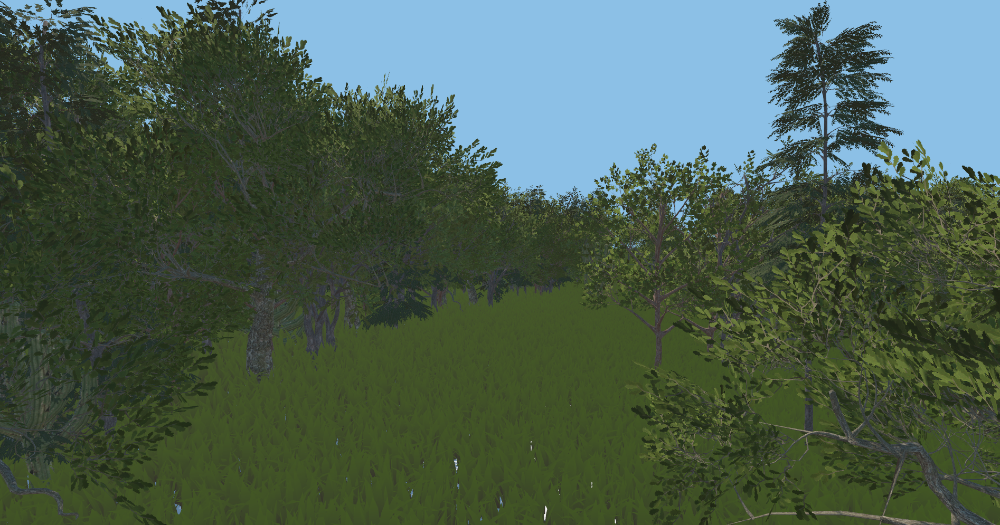}
    \end{subfigure}
\caption{Above: Frontal view of a generated forest sample scene. Below: Close up view.} \label{forestSceneUnityClose}
\end{figure}

\begin{figure}[t]
    \centering
    \begin{subfigure}[b]{0.15\textwidth}
        \includegraphics[width=\textwidth]{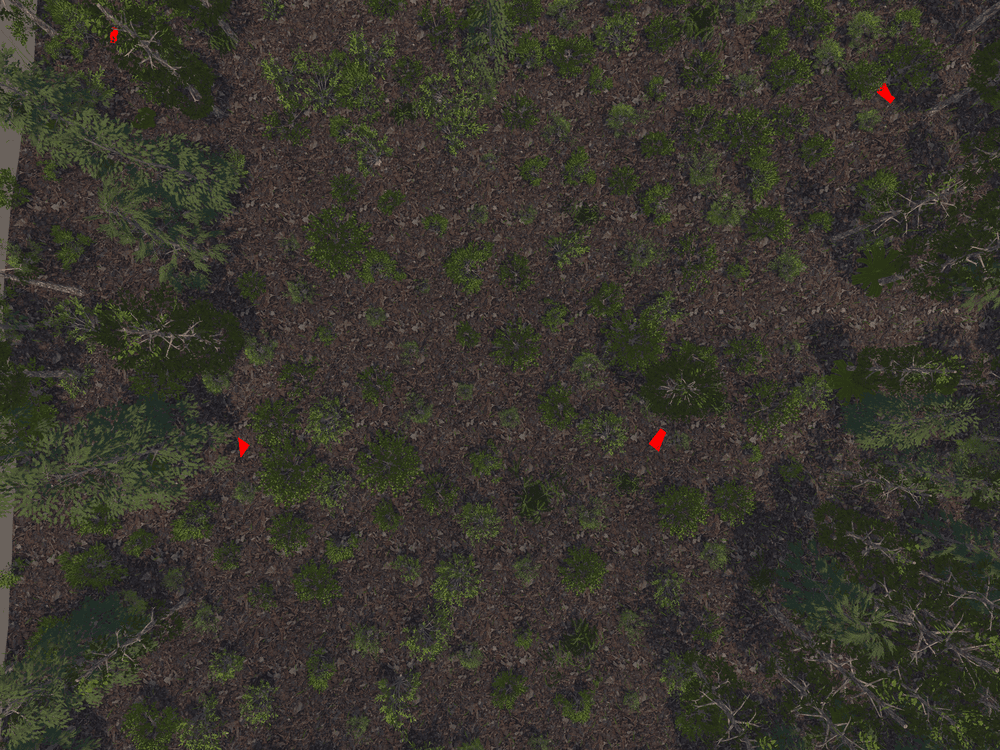}
    \end{subfigure}
    \begin{subfigure}[b]{0.15\textwidth}
        \includegraphics[width=\textwidth]{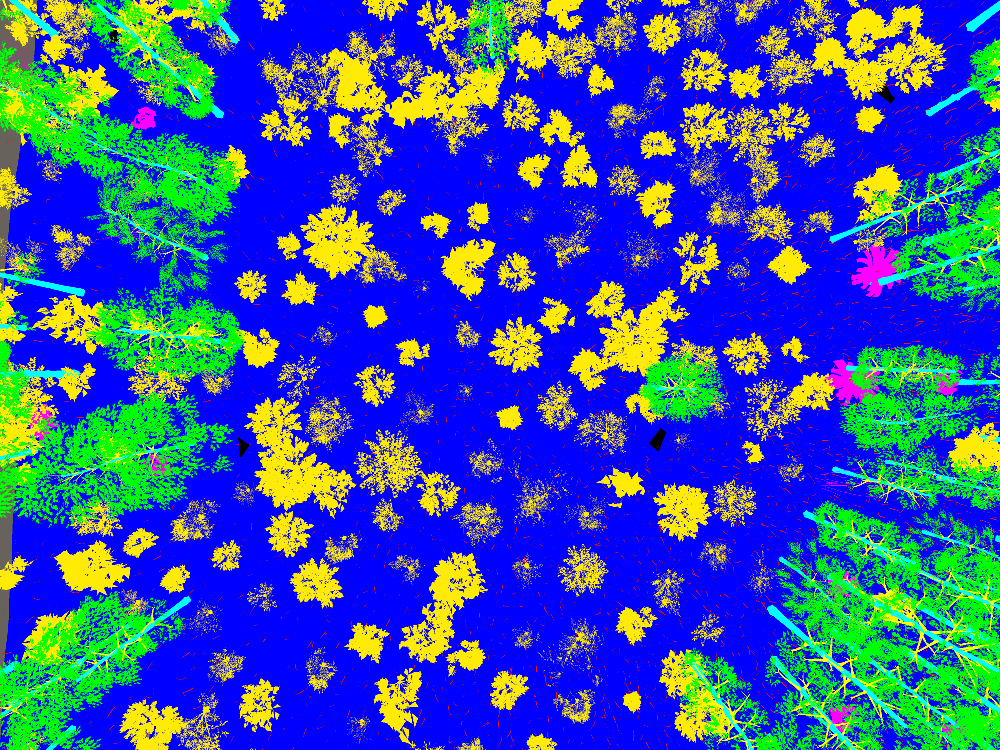}
    \end{subfigure}
    \begin{subfigure}[b]{0.15\textwidth}
        \includegraphics[width=\textwidth]{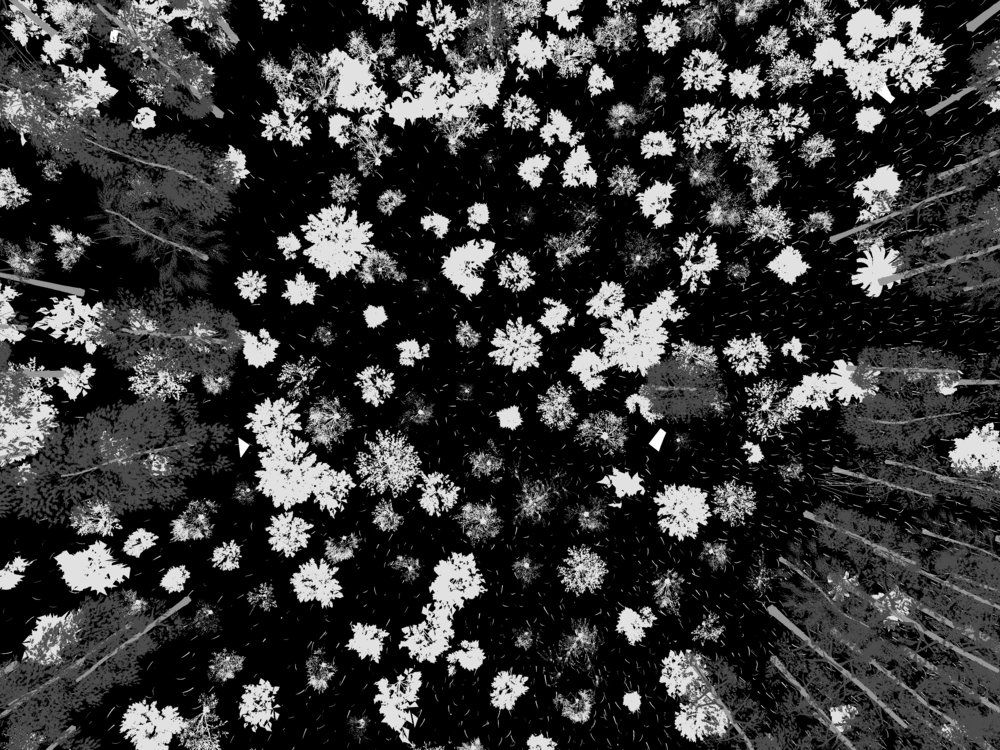}
    \end{subfigure} \\
    \vspace{1mm}
    \begin{subfigure}[b]{0.15\textwidth}
        \includegraphics[width=\textwidth]{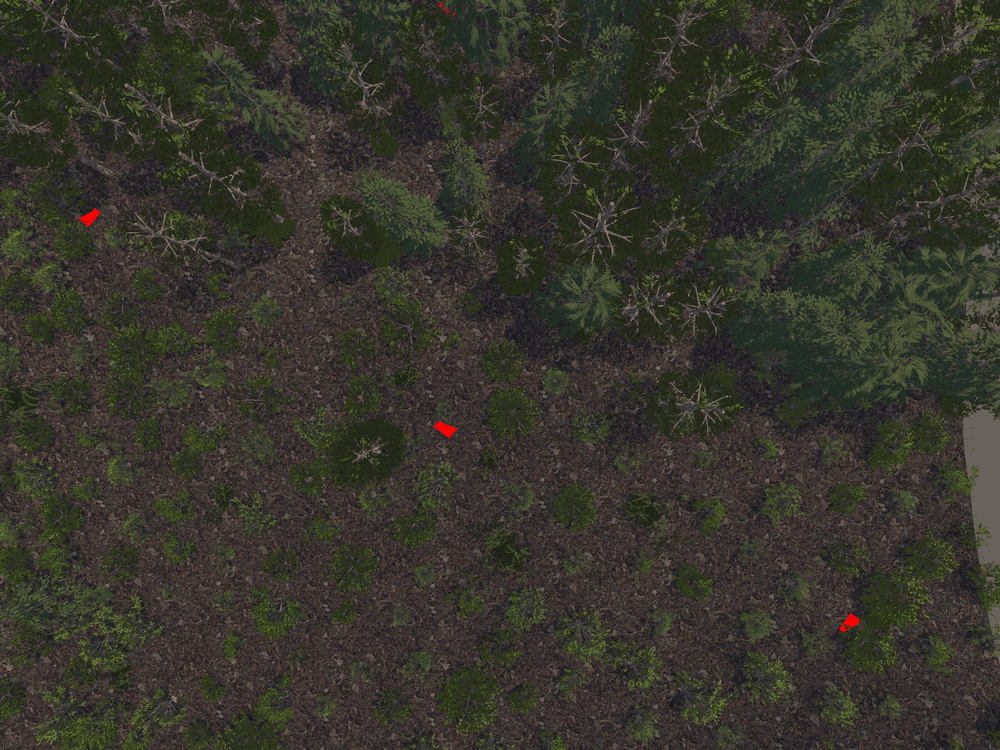}
    \end{subfigure}
    \begin{subfigure}[b]{0.15\textwidth}
        \includegraphics[width=\textwidth]{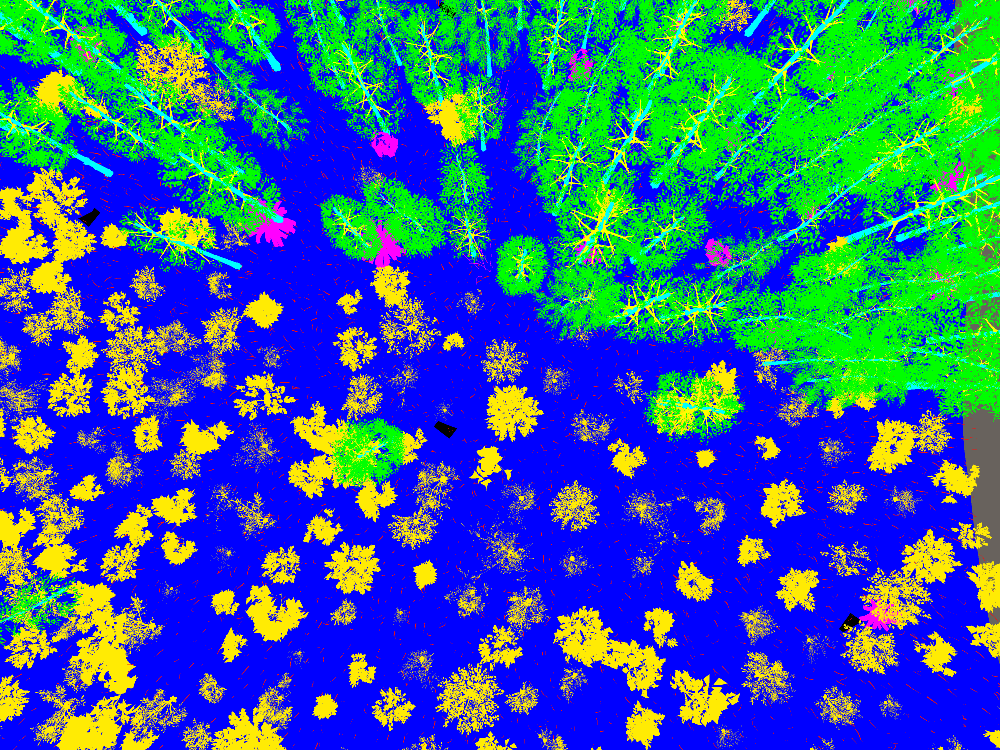}
    \end{subfigure}
    \begin{subfigure}[b]{0.15\textwidth}
        \includegraphics[width=\textwidth]{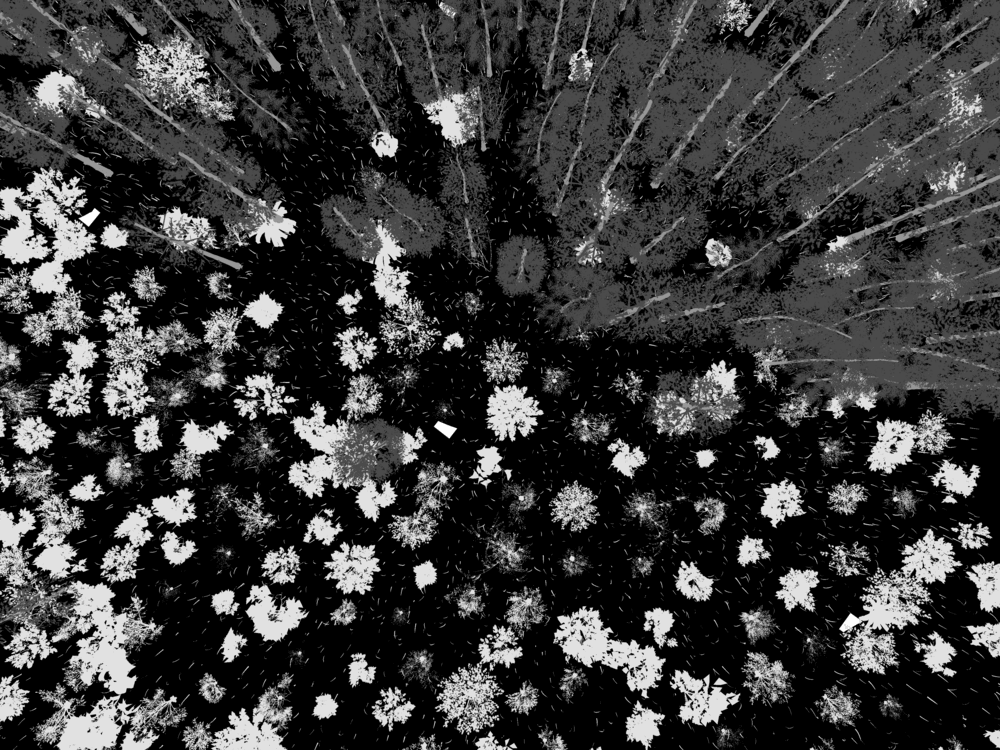}
    \end{subfigure} \\
    \vspace{1mm}
    \begin{subfigure}[b]{0.15\textwidth}
        \includegraphics[width=\textwidth]{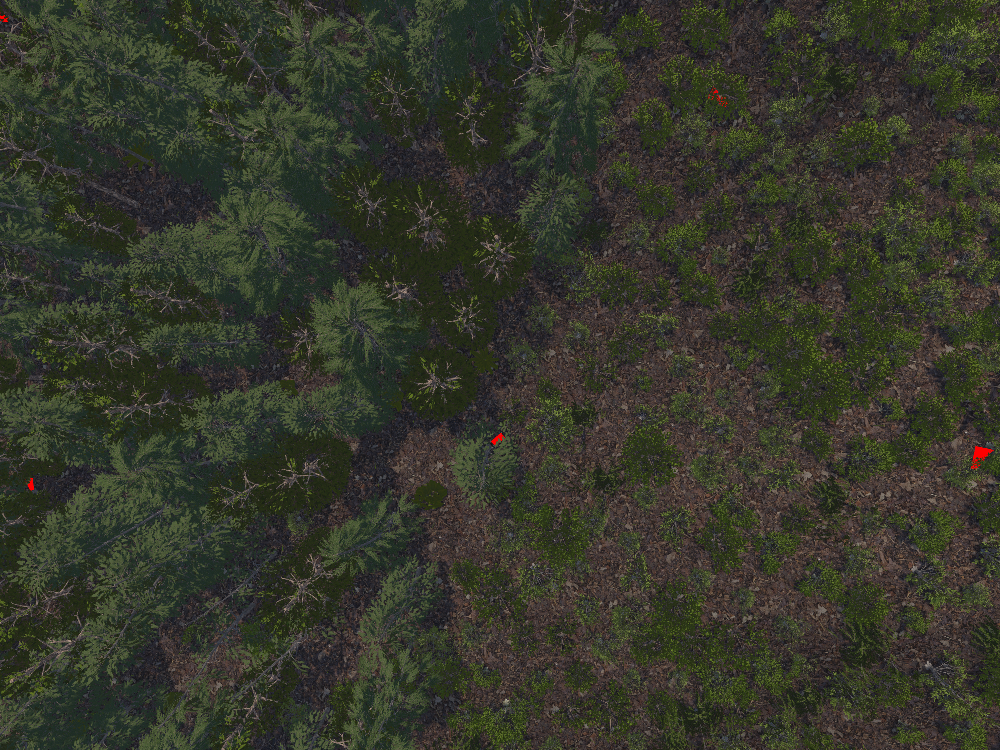}
    \end{subfigure}
    \begin{subfigure}[b]{0.15\textwidth}
        \includegraphics[width=\textwidth]{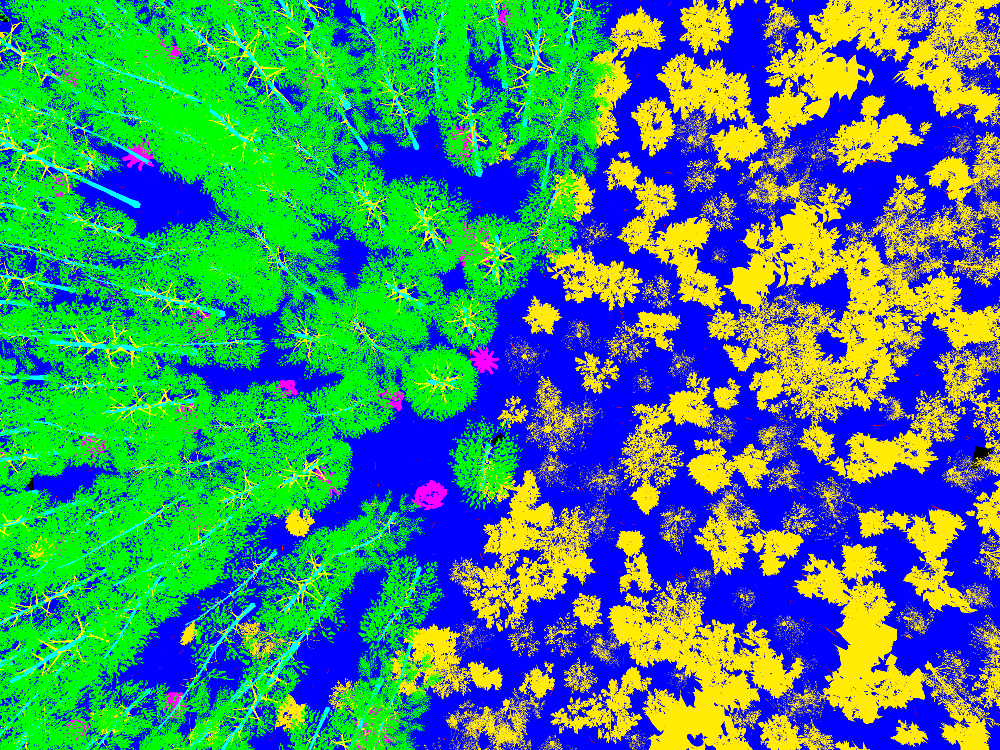}
    \end{subfigure}
    \begin{subfigure}[b]{0.15\textwidth}
        \includegraphics[width=\textwidth]{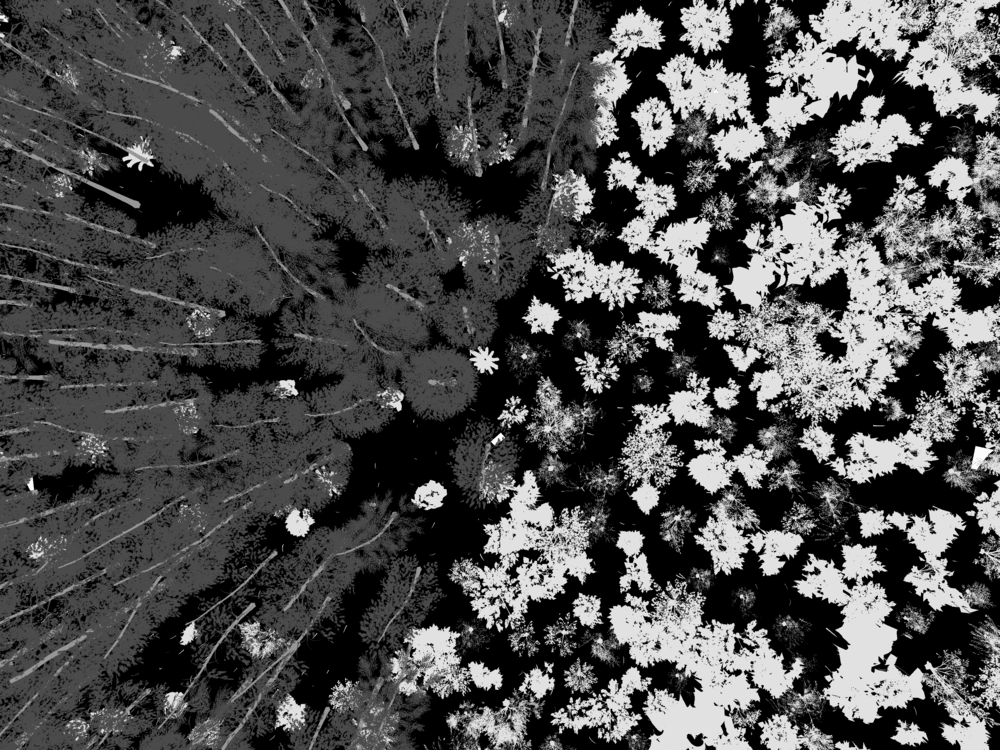}
    \end{subfigure}
\caption{Samples from the imagery taken from the forest simulator. Left: RGB images of the generated forest scene; the red dots corresponds to the ground control points. Middle: Corresponding semantic segmentation image; each color is related to a different category: tree canopies (green), tree trunks (cyan), bushes, grass and understorey vegetation (yellow), ground (blue) and ground control points (red). Right: Greyscale version of the semantic segmentation image, for its use in the SfM algorithm.} 
\label{forest-simulator-imagery}
\end{figure}

An example of a scene generated with this forest simulator can be seen in Fig.~\ref{forestSceneUnityClose}, and some of the top-view images obtained can be seen in Fig.~\ref{forest-simulator-imagery}. The last images are part of a small dataset of 45 images that will be used to test the implemented algorithms. The simulated survey was carried out at a flight altitude of 100 meters, and the photos were taken with an overlap of 85\% vertically and 80\% horizontally. The scene generation parameters were maintained the same as the ones used in~\cite{raverta2024}.

\subsection{Semantic Structure from Motion}

In this work, we build on top of OpenDroneMap, as it is a state-of-the-art, publicly available open-source Structure from Motion software for aerial photogrammetry. OpenDroneMaps includes desirable features, such as the possibility to add ground control points to get a more accurate reconstruction. This is a key feature for reconstructing difficult environments, as they consist of points with very high-precision geolocalization. OpenDroneMaps is made up of several modules, including OpenSfM~\cite{opensfm}, which performs the matching phase over every pair of images and estimates the camera pose and a sparse point cloud, OpenMVS~\cite{openmvs}, which performs the densification of the point cloud, and the FPCFilter module, which performs a fast point cloud filtering over the densified point cloud. 

Next, we list the modifications made to the OpenDroneMaps system, specifically to the three modules mentioned. The code and images used will be shared in a publicly available repository\footnote[2]{\url{https://github.com/lrse/SODM}}.

\subsubsection{Outlier Rejection}
In the OpenSfM module, during the matching stage between all pairs of images, an outlier rejection is added between matches that do not share the same semantic segmentation label. In this manner, we eliminate potential mismatches that might otherwise degrade the accuracy of the 3D reconstruction in the SfM process. Then, for each matched features $\left( f_i^k, f_j^l \right)$ between images $I_i$ and $I_j$ respectively, we only keep the matches that satisfy:
\begin{equation}
\mathcal{M}_{ij}^{\text{filtered}} = \left\{ \left( f_i^k, f_j^l \right) \in \mathcal{M}_{ij} \ \big| \ \textit{label}\left(f_i^k \right)= \textit{label}\left(f_j^l \right) \right\}
\end{equation}
where $\mathcal{M}_{ij}$ is the set of matches between images $I_i$ and $I_j$.

If only less than 15\% of the matches are considered inliers between two images, then the outlier rejection is not taken into account for that image pair, and all of the matches are conserved.

\subsubsection{Label and confidence estimation for each 3D point} 
During the point cloud densification stage in the OpenMVS module, since each three-dimensional point is tracked from several images, we retrieve the label of each of those images and keep the mode value. In this way, we assure that the most observed label from each of the possible views of that point is maintained, minimizing the possibility of mislabeling the 3D point $x$:
\begin{equation}
    label(x) = \mathop{\mathrm{mode}}\limits_{i \in I_x} [label(x_{proj_i})]
\end{equation}
where $I_x$ is the subset of images where the 3D point $x$ is viewed and tracked, and $label(x_{proj_i})$ is the label of the projection of the 3D point $x$ on image $i$.

Additionally, the confidence of each of the labels was estimated by calculating the proportion of the labels of the projected points in the images that coincide with the resulting mode out of all the images on the subset:
\begin{equation}
    confidence(x) = \frac{\left| \left\{ i \in I_x \;\middle|\; \textit{label}(x_{\text{proj}_i}) = \textit{label}(x) \right\} \right|
}{|I_x|}
\end{equation}

It can be seen that as the confidence value gets closer to 1, more of the 3D point projections on the images that track its position will share the same semantic segmentation value. The FPCFilter module of OpenDroneMaps was adapted to receive point clouds with these modifications accordingly.

\subsubsection{Export of the label and confidence values}
The estimated values for the labels and confidence of each point of the resulting point cloud are exported in a \textit{.ply} file, along with the coordinates of the point and its associated RGB color. 

\section{Results and Discussion}

\begin{figure}[t]
\centerline{\includegraphics[width=.45\textwidth]{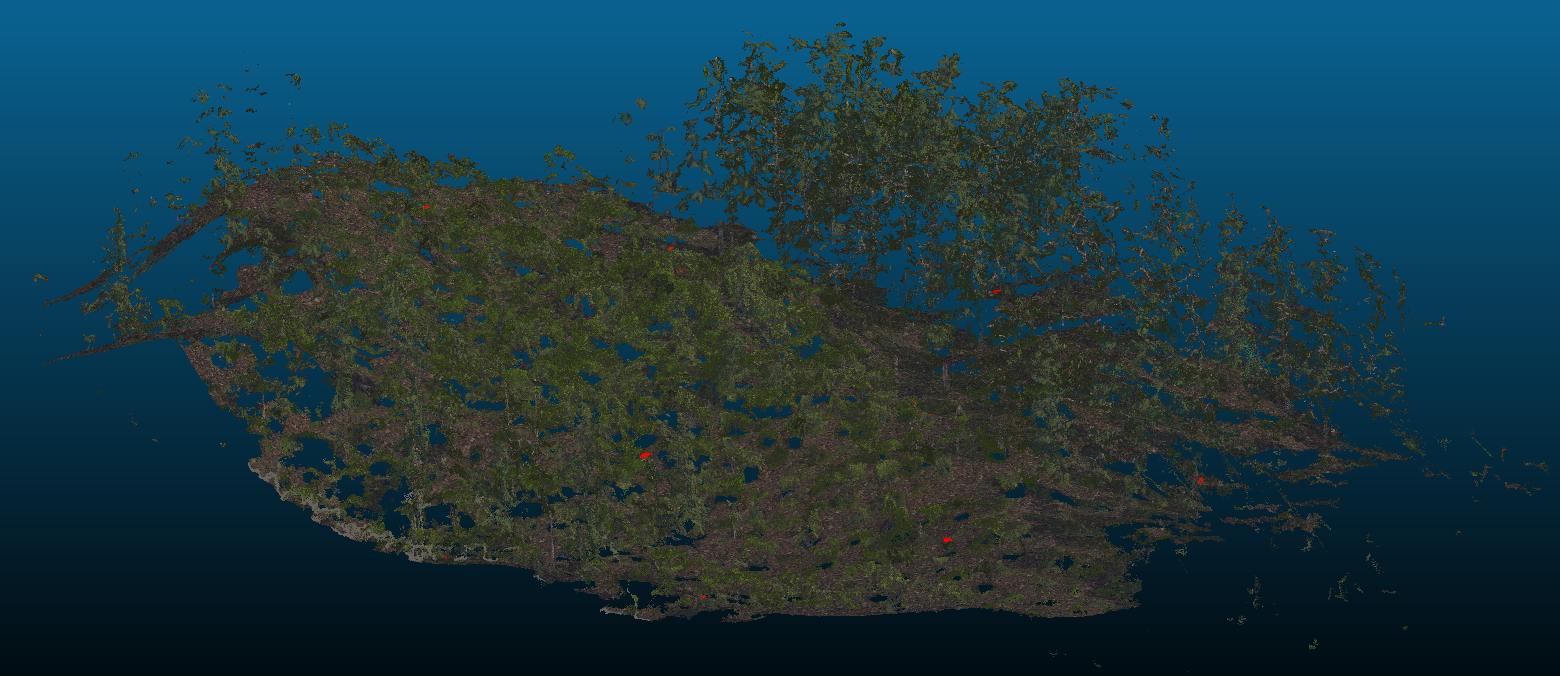}}
\caption{Point cloud obtained using the modified OpenDroneMaps, with the respective RGB color of each point.}
\label{sfm-results}
\end{figure}

\begin{figure}[t]
\centerline{\includegraphics[width=.45\textwidth]{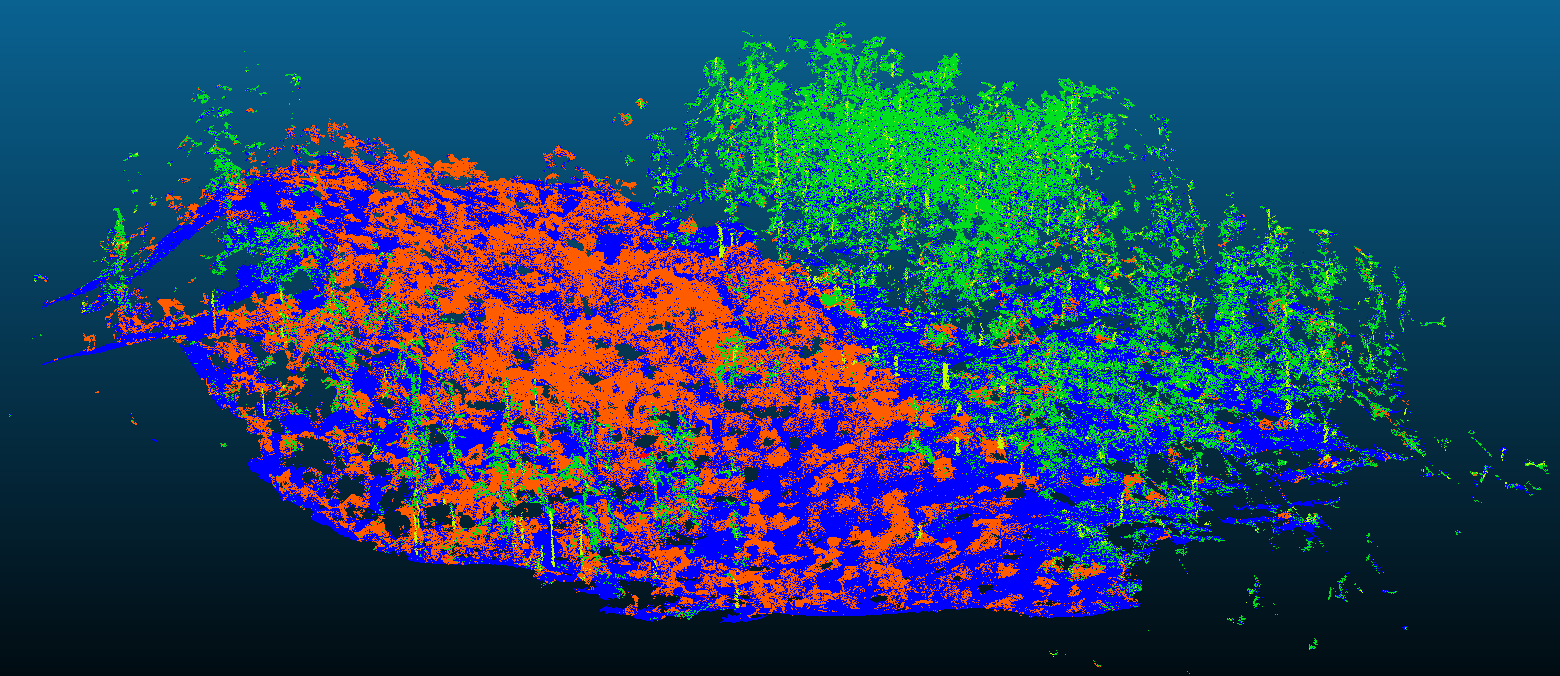}}
\caption{Semantic segmentation obtained for the point cloud of the synthetic forest scene. Each color represent a different category: ground (blue), tree trunks (yellow), tree canopies (green) and bushes, grass and understorey (orange).}
\label{segmentation-results}
\end{figure}

\begin{figure}[t!]
    \centering
    \begin{subfigure}[b]{0.24\textwidth}
        \includegraphics[width=\textwidth]{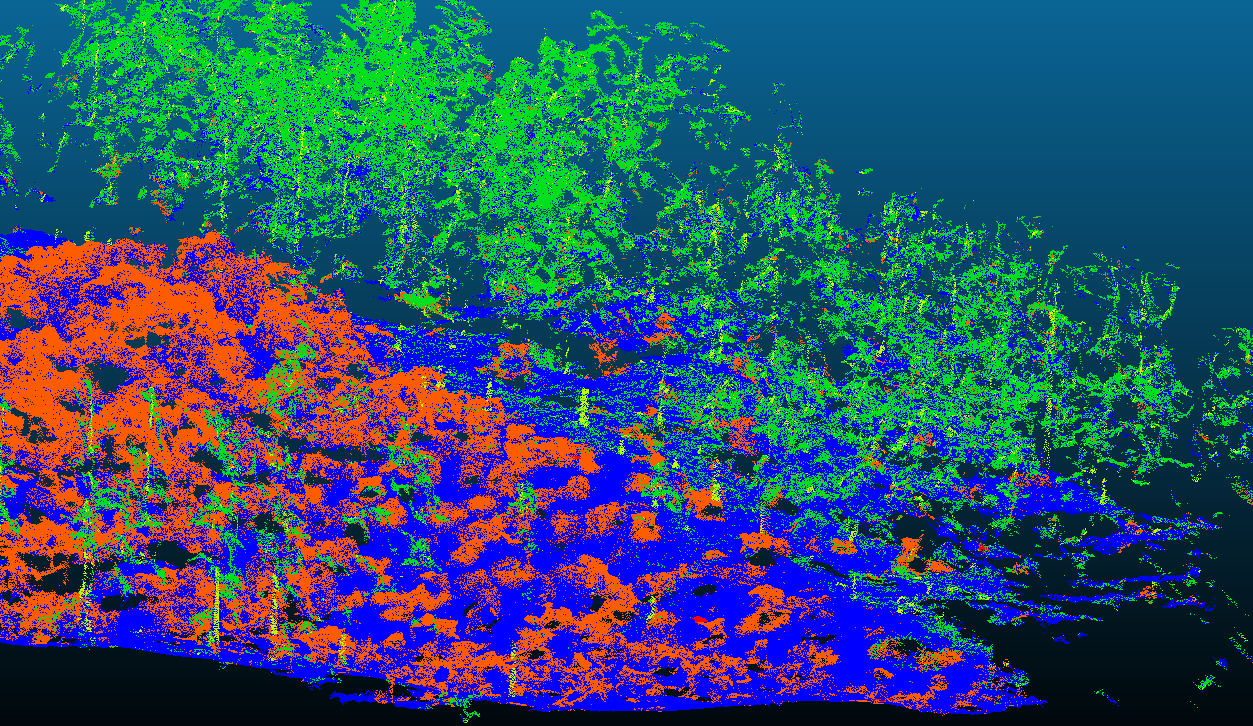}
    \end{subfigure}
    \begin{subfigure}[b]{0.24\textwidth}
        \includegraphics[width=\textwidth]{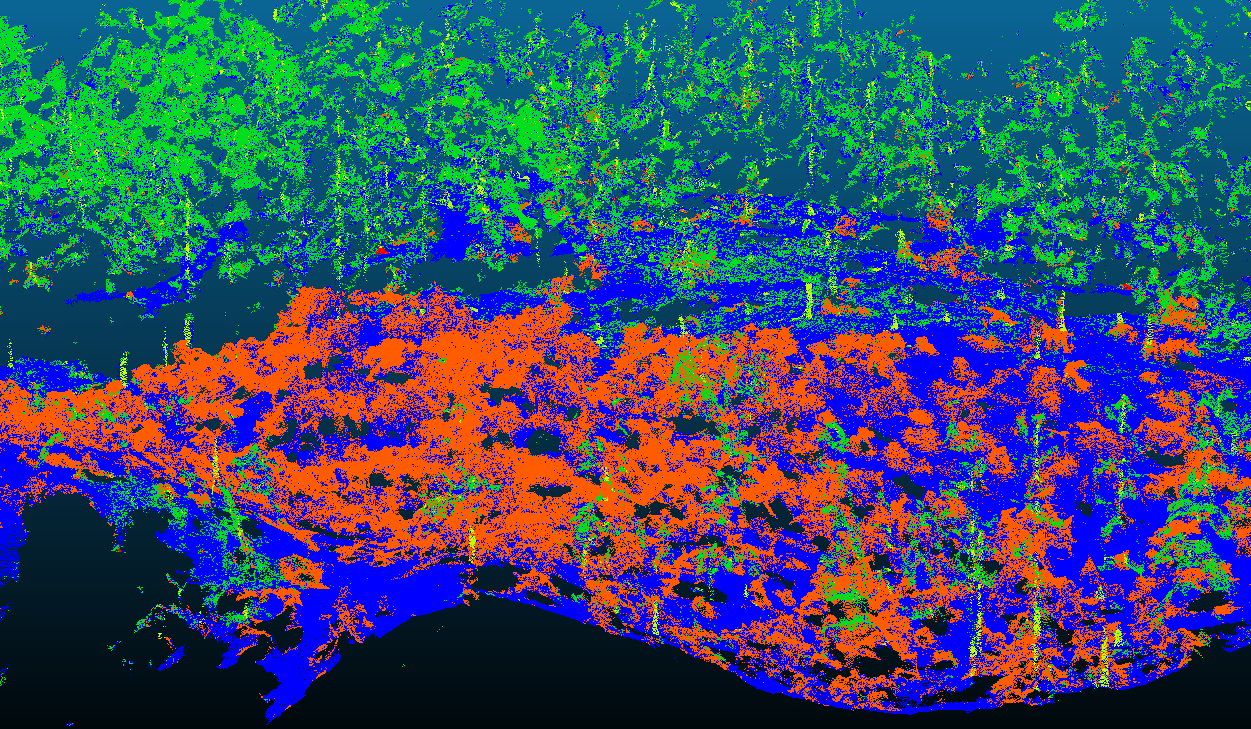}
    \end{subfigure}
\caption{Close up view from two different perspectives of the semantic segmentation obtained. Each color represent a different category: ground (blue), tree trunks (yellow), tree canopies (green) and bushes, grass and understorey (orange).} \label{segmentation-results-close-up}
\end{figure}

\begin{figure}[b]
\centerline{\includegraphics[width=.45\textwidth]{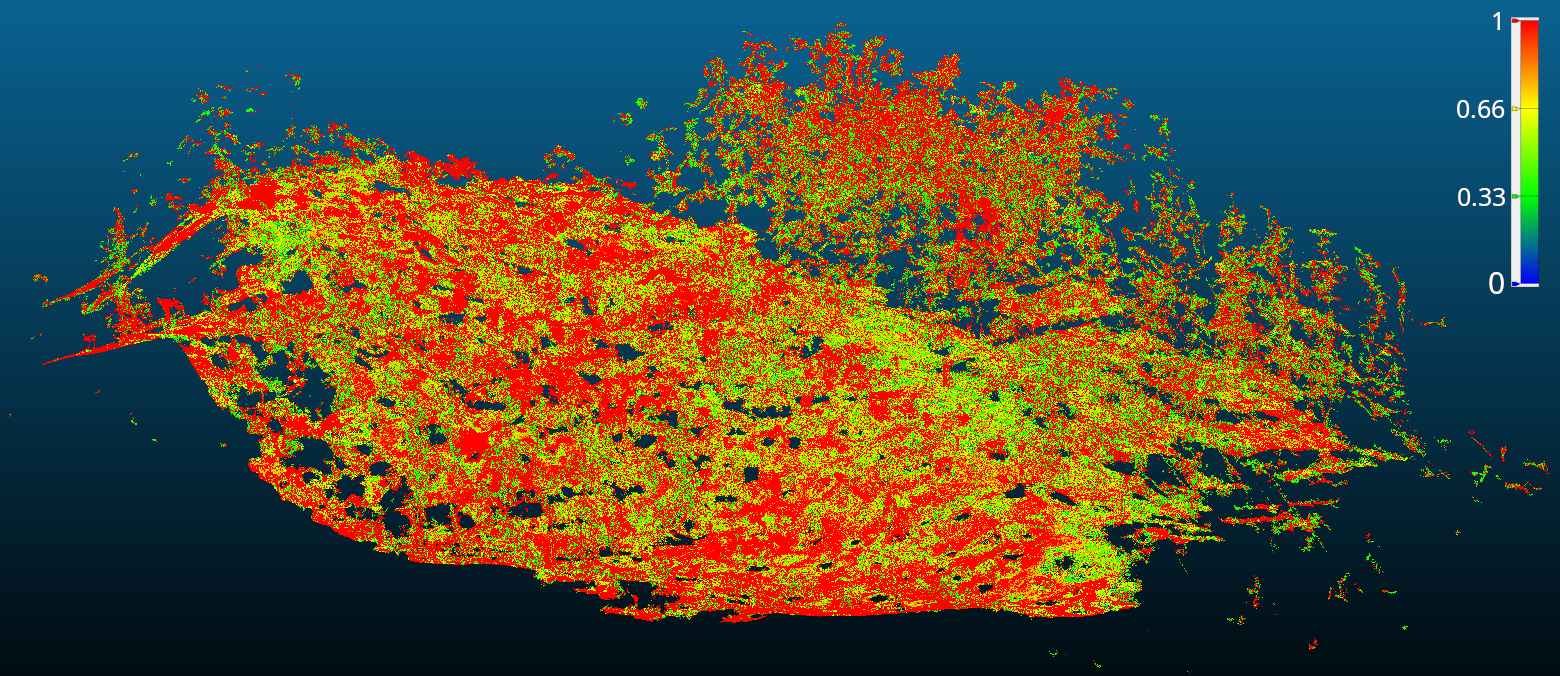}}
\caption{Confidence of the semantic segmentation obtained for the point cloud of the synthetic forest scene.}
\label{confidence-results}
\end{figure}

This section shows the results obtained with the 45 input images and their respective semantic segmentation masks, some of which are shown in Fig.~\ref{forest-simulator-imagery}. In Fig.~\ref{sfm-results} the results of applying the modified OpenDroneMaps to the set of images can be seen. Figs.~\ref{segmentation-results} and~\ref{segmentation-results-close-up} show the estimated segmentation obtained, and~\ref{confidence-results} show the segmentation confidence estimation. It can be noticed that the segmentation is consistent with the semantic segmentation shown in Fig.~\ref{forest-simulator-imagery}, distinguishing between tree canopies, tree trunks, bushes and understorey vegetation and ground. Also, we can see that the confidence shows good results overall, but the best results are in pixels located in places where there are no noticeable height differences compared to neighboring points. This occurs because reprojection errors of the 3D point onto the images often lead to incorrect segmentation values, particularly near category boundaries, such as between the ground and tree foliage, as another segmentation value is likely to be found. 

\begin{figure}[h]
\centerline{\includegraphics[width=.45\textwidth]{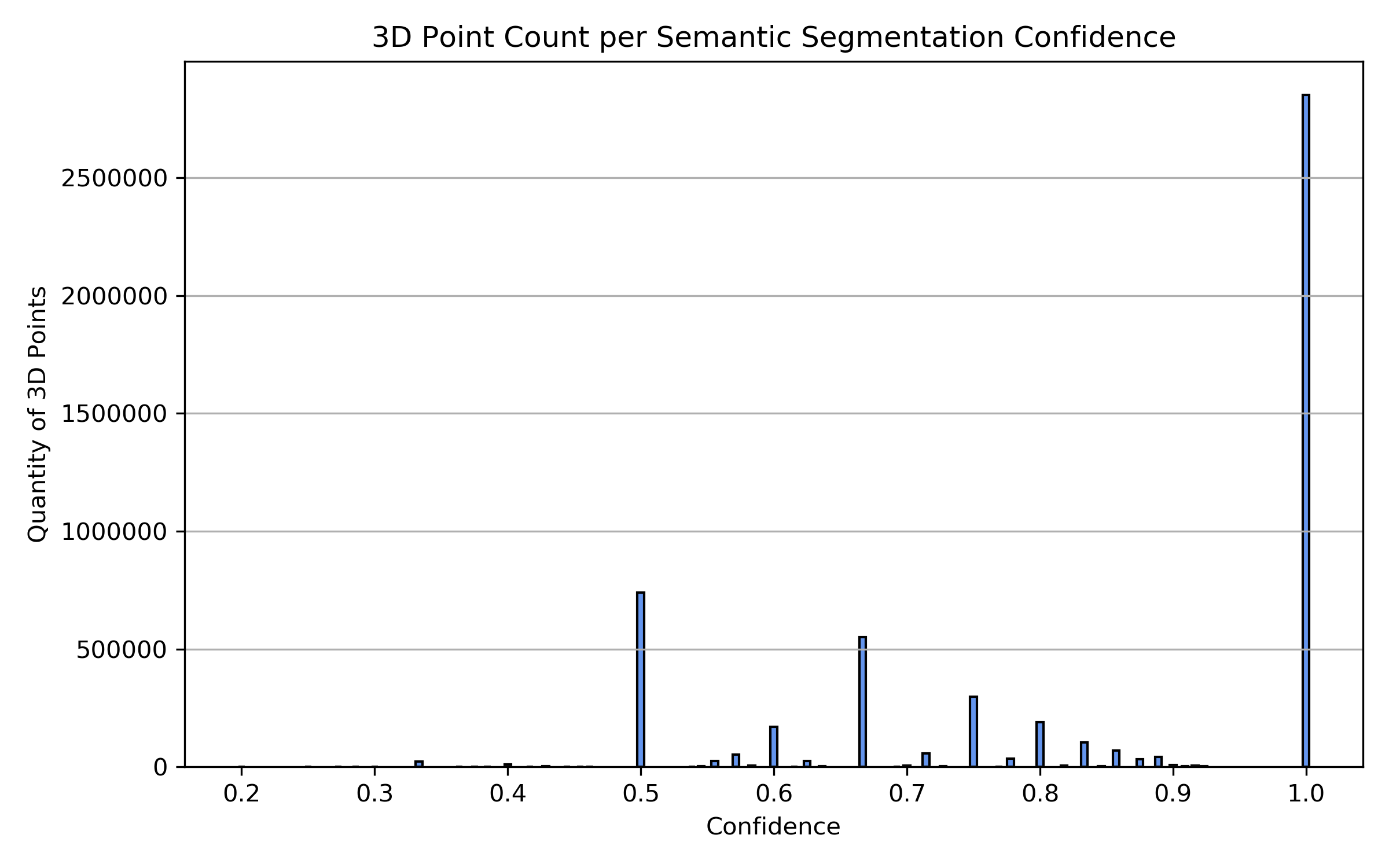}}
\caption{Resulting 3D point count per confidence obtained. The closer the number is to 1, the more confident the algorithm is in the estimated value of the semantic segmentation label.}
\label{histogramConfidence}
\end{figure}

In Fig.~\ref{histogramConfidence} a bar plot of the estimated confidence can be seen, which shows that the majority of points (about 54\%) have confidence equal to 1, indicating that the same semantic segmentation label was seen in all images that tracked that point. Depending on the importance of the semantic segmentation labels for a given application, a filter may be used to retain only the points with higher confidence.

\section{Conclusions and Future Work}
\subsection{Conclusions}
In this work, we developed an open-source modification of the OpenDroneMaps software to increase its functionality by using the semantic segmentation information of the input images to infer the corresponding labels of each point of the dense point cloud given as output. Additionally, this information was used to filter outliers in the matching phase and thus reach a more accurate 3D representation of the reconstructed synthetic forest scene studied. 

\subsection{Ongoing work}
We are studying possible ways to use the input semantic segmentation data to achieve better results in the overall reconstruction of the forest scene, by modifying the OpenDroneMaps system beyond the matching phase, such as the Constrained Bundle Adjustment Algorithm proposed by~\cite{chen2018}.
In addition, the modifications implemented will be tested using real forest data with semantic segmentation information as input, such as that presented in~\cite{cloutier2023}. 

\subsection{Future work}
As future work, we plan to extend this processing pipeline by training deep learning architectures capable of segmenting point clouds with the point clouds obtained with the methodology explained in this work.  These networks will then be evaluated on new 3D forest reconstructions captured through photogrammetry in real-world environments. Our final goal is to develop models capable of accurately segmenting image-derived forest point clouds into meaningful categories, thereby enabling a more detailed analysis of the studied ecosystems.






\end{document}